\documentclass[letterpaper]{article} 
\usepackage{aaai24}  
\usepackage{times}  
\usepackage{helvet}  
\usepackage{courier}  
\usepackage[hyphens]{url}  
\usepackage{graphicx} 
\usepackage{natbib}  
\usepackage{caption} 
\usepackage{algorithm}
\usepackage{algorithmic}
\usepackage{subfigure}
\usepackage{multirow,makecell}
\usepackage{amsmath}
\usepackage{amsfonts}
\usepackage{booktabs}
\usepackage{tabularx}
\usepackage{caption}
\usepackage[page]{appendix} 
\usepackage{newfloat}
\usepackage{listings}
\DeclareCaptionStyle{ruled}{labelfont=normalfont,labelsep=colon,strut=off} 
\floatstyle{ruled}
\newfloat{listing}{tb}{lst}{}
\floatname{listing}{Listing}
\nocopyright 

\title{V2A-Mapper: A Lightweight Solution for Vision-to-Audio Generation by Connecting Foundation Models}
\author {
    Heng Wang\textsuperscript{\rm 1}\thanks{Work done during an internship at Dolby.},
    Jianbo Ma\textsuperscript{\rm 2},
    Santiago Pascual\textsuperscript{\rm 2},
    Richard Cartwright\textsuperscript{\rm 2},
    Weidong Cai\textsuperscript{\rm 1}
}
\affiliations {
    \textsuperscript{\rm 1}University of Sydney\\
    \textsuperscript{\rm 2}Dolby Laboratories\\
    \{heng.wang, tom.cai\}@sydney.edu.au, \{jianbo.ma, santiago.pascual, richard.cartwright\}@dolby.com
    
    {\normalsize\url{https://v2a-mapper.github.io/}}
}

\usepackage{bibentry}
\begin{document}
\maketitle

\begin{abstract}
Building artificial intelligence (AI) systems on top of a set of foundation models (FMs) is becoming a new paradigm in AI research. Their representative and generative abilities learnt from vast amounts of data can be easily adapted and transferred to a wide range of downstream tasks without extra training from scratch. However, leveraging FMs in cross-modal generation remains under-researched when audio modality is involved. On the other hand, automatically generating semantically-relevant sound from visual input is an important problem in cross-modal generation studies. To solve this vision-to-audio (V2A) generation problem, existing methods tend to design and build complex systems from scratch using modestly sized datasets. In this paper, we propose a lightweight solution to this problem by leveraging foundation models, specifically CLIP, CLAP, and AudioLDM. We first investigate the domain gap between the latent space of the visual CLIP and the auditory CLAP models. Then we propose a simple yet effective mapper mechanism (V2A-Mapper) to bridge the domain gap by translating the visual input between CLIP and CLAP spaces. Conditioned on the translated CLAP embedding, pretrained audio generative FM AudioLDM is adopted to produce high-fidelity and visually-aligned sound. Compared to previous approaches, our method only requires a quick training of the V2A-Mapper. We further analyze and conduct extensive experiments on the choice of the V2A-Mapper and show that a generative mapper is better at fidelity and variability (FD) while a regression mapper is slightly better at relevance (CS). Both objective and subjective evaluation on two V2A datasets demonstrate the superiority of our proposed method compared to current state-of-the-art approaches - trained with 86\% fewer parameters but achieving 53\% and 19\% improvement in FD and CS, respectively. Supplementary materials such as audio samples are provided at our demo website: \url{https://v2a-mapper.github.io/}.
\end{abstract}

\begin{figure}[!t]
\centering     
\subfigure[Previous V2A generation approaches.]{\includegraphics[width=0.4\textwidth]{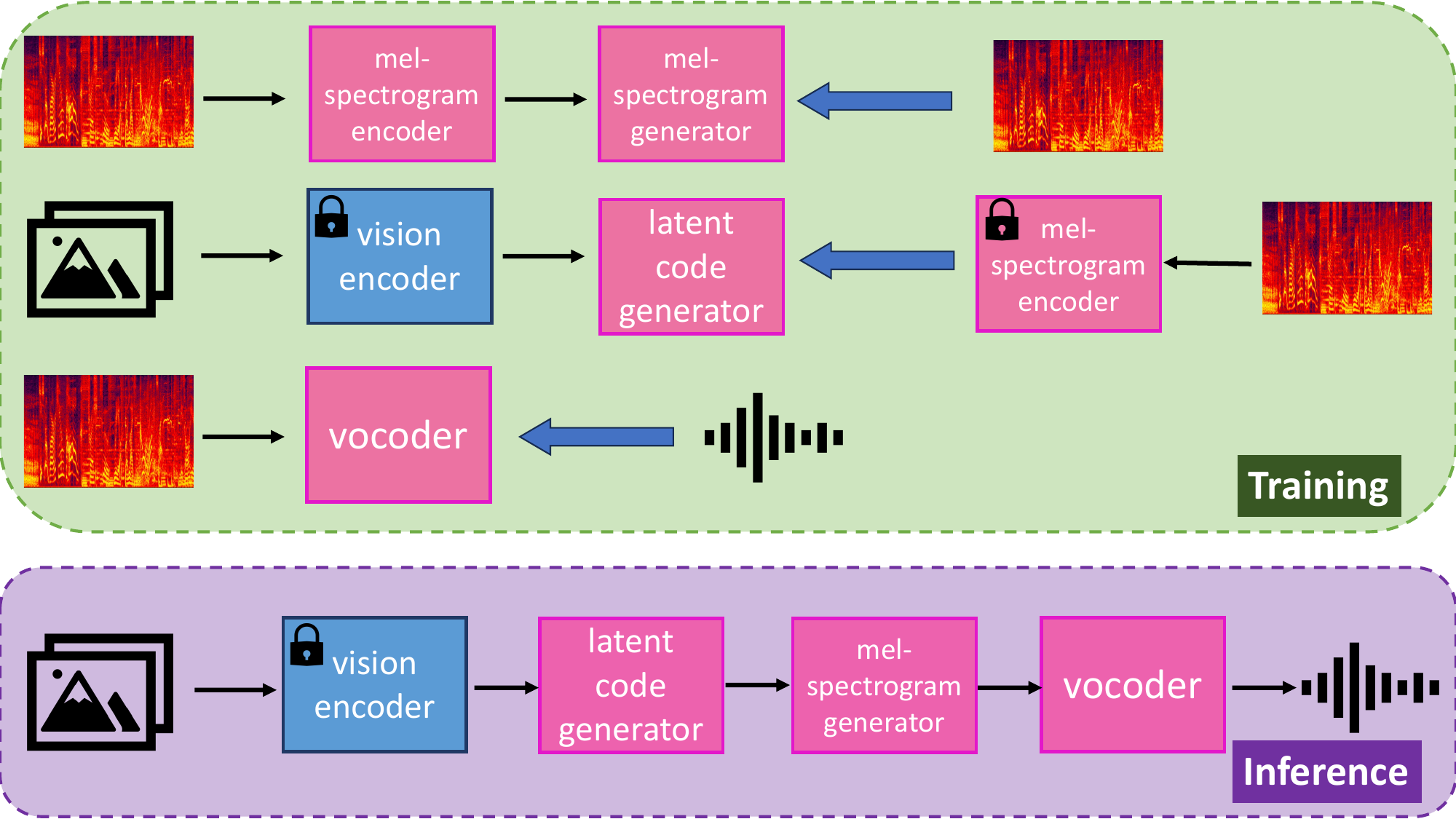}\label{fig:teasera}}
\subfigure[Our lightweight solution to V2A generation.]{\includegraphics[width=0.4\textwidth]{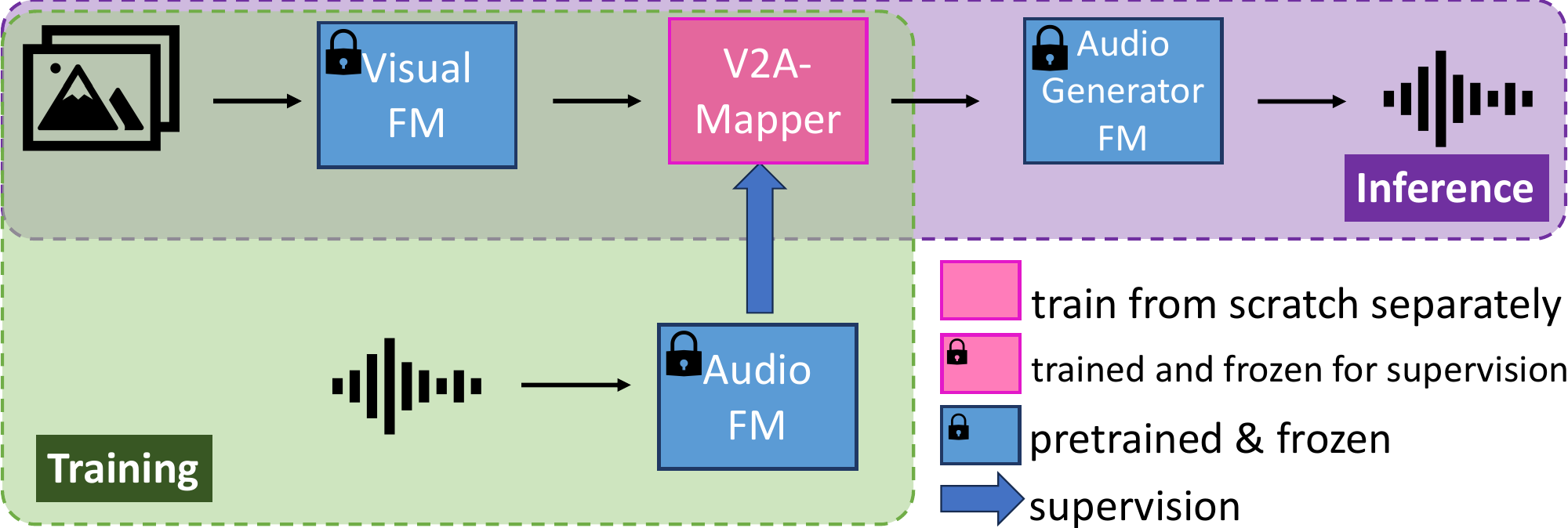}\label{fig:teaserb}}
\caption{Schematic illustrations of training and inference pipelines from previous V2A algorithms and our lightweight solution, respectively. Leveraging foundation models (FMs), we only require the training of a single V2A-Mapper while current works involve multiple modules to train.}
\label{fig:teaser}
\end{figure}
\section{Introduction}
Foundation models (FMs), trained on large-scale data and often making use of self-supervised learning, offer problem-agnostic representative or generative capabilities to downstream tasks via adaptation~\cite{fm}. They have demonstrated robust generalization and knowledge transfer ability across a broad spectrum of tasks in recent AI research~\cite{fm_survey1,fm_survey2,fm_survey3}. Despite success in many uni-modal tasks spanning language~\cite{paass2023foundation}, vision~\cite{awais2023foundational}, and audio~\cite{fm_adap_audio}, the adaptation of FMs in problems involving multiple modalities such as cross-modal generation is greatly dominated by vision-language research~\cite{du2022survey}. Although attempts~\cite{ao2022speecht5, audiogen,diffsound,makeanaudio,audioldm,audioldm2,leverage_auidoldm_soundg,tango} have been made lately to bring FMs into text-to-audio generation and achieved remarkable performance, the viability of adopting FMs in vision-to-audio generation is still unclear. 

Vision and audio are two essential and correlated sources through which people perceive the world. Humans have the ability to imagine the corresponding sound when just observing a visual event~\cite{owens2018audio}. Mimicry of this human-like cross-modal generation ability is applicable to various scenarios such as enhancing the experience of immersion in virtual reality, automating video editing for content creators, and assisting people with visual impairment~\cite{autofoley,difffoley}. Such rich visual-audio consistency and wide application have drawn constant interest in vision-to-audio (V2A) generation~\cite{av_learning}. Not restricted to a specific in-domain sound type (e.g., background music~\cite{di2021video}, dance music~\cite{zhu2022discrete}, speech~\cite{prajwal2020learning}), in this paper, we aim to generate natural sound from visual input in more diverse real-world scenarios, a V2A task that poses a markedly elevated level of difficulty~\cite{vegas}. 

To solve this open-domain V2A generation problem, current methods~\cite{specvqgan,im2wav,clipsonic} often involve a complex system of separately optimized submodules trained with limited size of datasets as illustrated in Fig.~\ref{fig:teasera}. It could be cumbersome and resource-intensive to train each module individually and the generalization capability of each module could be restricted due to the lack of sufficient training data.

In this work, we explore the feasibility of adopting foundation models in open-domain vision-to-audio generation task. As shown in Fig.~\ref{fig:teaserb}, our lightweight method only requires the training of a \textbf{V2A-Mapper} to bridge the domain gap between the vision representative FM CLIP~\cite{clip} and the audio generative FM AudioLDM~\cite{audioldm}. The V2A-Mapper is supervised by the audio representative FM CLAP~\cite{clap} to learn the translation from visual space to auditory space. Leveraging the generalization and knowledge transfer ability of foundation models, the V2A-Mapper is trained with the same modestly sized dataset but the overall system can achieve much better performance. Our contribution includes: \textbf{1)} investigating the potential of bringing FMs into the field of vision-to-audio generation; \textbf{2)} proposing a simple but effective V2A-Mapper to connect visual and auditory FMs; \textbf{3)} investigating both generative and regression strategies of the V2A-Mapper; \textbf{4)} both subjective and objective evaluation on two V2A datasets demonstrate the efficiency and effectiveness of our method - it is trained with 86\% fewer parameters but can achieve up to 53\% and 19\% improvement in fidelity (FD) and relevance (CS).

\section{Related Works}
\subsubsection{Vision-to-Audio Generation.}
Earlier V2A works~\cite{owens2016visually,chen2017deep,hao2018cmcgan} deal with limited sound in controlled environments. VEGAS~\cite{vegas} for the first time introduced open-domain sound generation from in-the-wild visual input. But VEGAS and later works~\cite{chen2018visually,chen2020generating} had to train a separate model for each sound type which is hard to scale up. To solve this issue, SpecVQGAN~\cite{specvqgan} designed the first label-free approach where a single model can produce diverse sound types. SpecVQGAN used a pretrained image classifier network to extract visual features from which a Transformer-based~\cite{transformer} autoregressive model synthesizes the mel-spectrogram. Upgrading this label-free approach, Im2Wav~\cite{im2wav} used the vision foundation model CLIP~\cite{clip} to get visual features of multimodal semantic information. Instead of predicting the mel-spectrogram directly, Im2Wav autoregressively generates its latent code based on the visual prompt and a VQ-VAE~\cite{vqvae} is trained to encode and decode between the mel-spectrogram and the latent space as shown in Fig.~\ref{fig:teasera}. Similar to Im2Wav, CLIPSonic-IQ~\cite{clipsonic} also adopted CLIP but they trained a diffusion model~\cite{clipsonic_ldm} to directly generate the mel-spectrogram as in SpecVQGAN. All of these attempts train multiple modules with limited amount of data from scratch. In this paper, we propose to utilize FMs to inherit their generalization ability obtained from large-scale training. Optimizing a V2A-Mapper to connect FMs with the same modestly sized dataset, our method is lightweight in the training phase and effective in the generalization capability. 
\begin{figure*}[!t]
\centering     
\includegraphics[width=\textwidth]{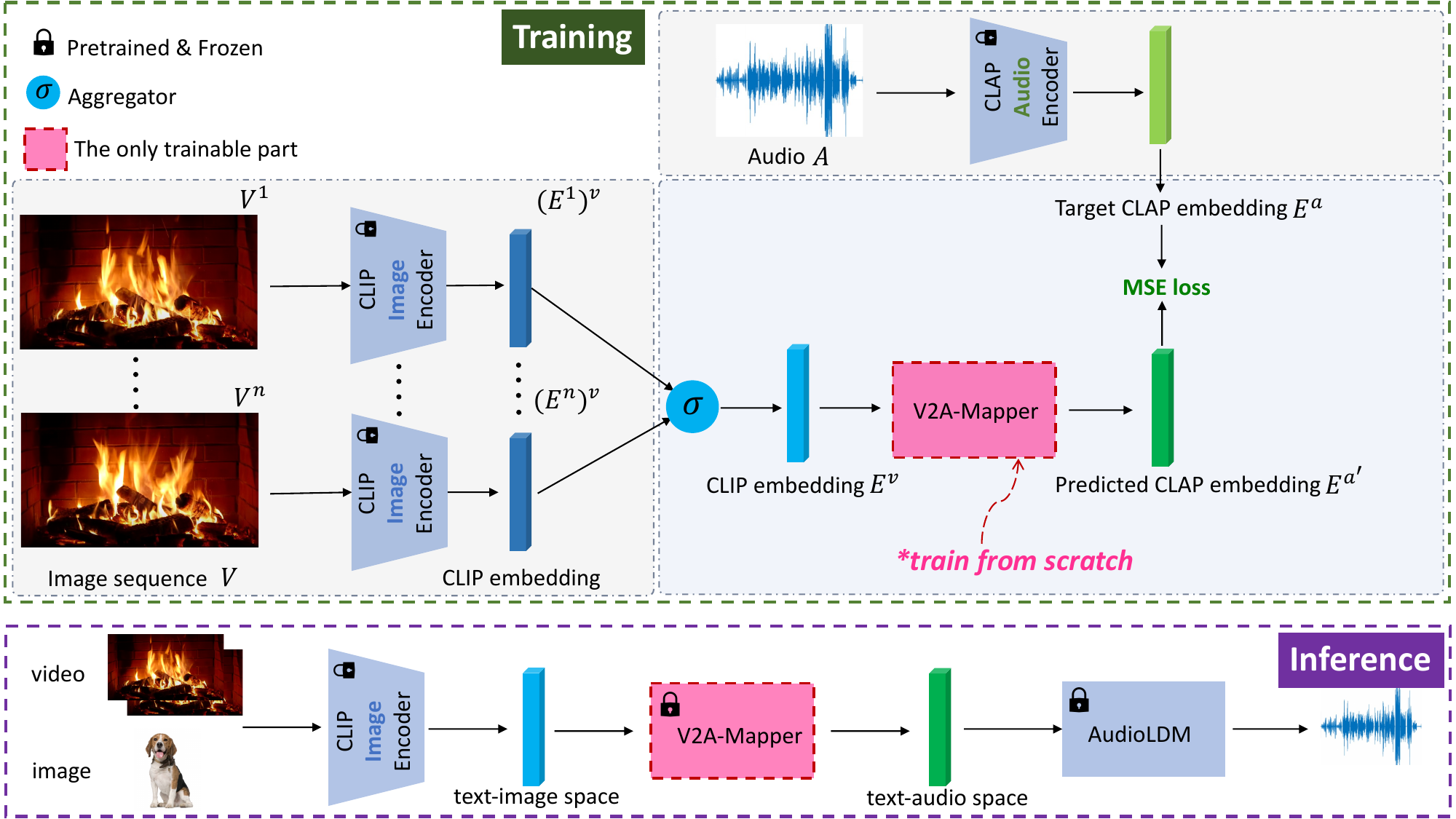}
\caption{\textbf{Top: Our lightweight V2A-Mapper training process.} We first extract the source visual embedding $E^v$ and the target audio embedding $E^a$ with frozen pretrained foundation models CLIP and CLAP. We explore different aggregators $\sigma$ to project the video data into a single feature vector. We then train the proposed V2A-Mapper using the audio-visual pair $\{E^v, E^a\}$ with MSE loss. \textbf{Bottom: The compact inference pipeline of our method for vision-to-audio generation.} We first adopt pretrained CLIP image encoder to project video/image into text-image space (the aggregation process for video input is omitted for brevity) and then use the trained V2A-Mapper to translate the visual embedding into CLAP text-audio space. Conditioned on the pseudo CLAP audio embedding, AudioLDM can be utilized to produce the sound waveform.}
\label{fig:overview}
\end{figure*}

\subsubsection{Foundation Model Adaptation.}
Adapting FMs to downstream tasks has been actively explored in NLP for uni-modal tasks, which can be categorized into prompt-based~\cite{prompt_based}, fine-tune-based~\cite{finetune_based, finetune_based_2}, and lightweight adapter-based methods~\cite{lightweight_adapt}. When introducing this new paradigm into multimodal domain, pioneering works in the vision-language (VL) field follow the third strategy and freeze FMs to avoid catastrophic forgetting~\cite{catastrophic_forgetting}. PICa~\cite{vl_adapt1} considered language FM GPT3 as a knowledge base for visual question answering tasks while ClipCap~\cite{vl_adapt2} and Flamingo~\cite{flamingo} learnt auxiliary modules (i.e., interleaving new layers or tokens) to utilize vision and language FMs for image captioning. Compared to VL field, there is much less research on FM adaptation in vision-audio domain. In this paper, we propose a simple yet effective V2A-Mapper to connect visual and auditory FMs for open-domain V2A generation task. In line with VL works, we keep our FMs frozen but, unlike previous attempts, we do not change the inner architecture of FMs. Our method keeps FMs completely intact and only adds a mapper, which guarantees easy deployment and updating.

\section{Method}
Our lightweight solution includes a visual encoder FM (CLIP), an audio encoder FM (CLAP), an audio generator FM (AudioLDM), and a trainable V2A-Mapper. Fig.~\ref{fig:overview} presents how we train the V2A-Mapper with frozen CLIP and CLAP models and how we incorporate it with frozen CLIP and AudioLDM models to produce high-fidelity and visually-aligned sound. In this section, we first revisit the adopted foundation models. We then analyze the domain gap between visual and auditory spaces and introduce how we train the V2A-Mapper to bridge the gap. Lastly, we present the details of our generative diffusion-based V2A-Mapper.

\subsection{Selected Foundation Models}
\label{sec:fm}
We choose the following foundation models because they are currently the state-of-the-art FMs for vision representation, audio representation, and audio generation, respectively. They can be replaced given better alternatives.
\subsubsection{CLIP.}
As our V2A generation task spans across two modalities, adapting multimodal FMs is a natural way of utilizing their semantic features for tasks involving multiple domains~\cite{lu2021pretrained}. CLIP~\cite{clip} is a text-image representation model which is trained to maximize the similarity between 400M paired text and image data via contrastive learning. Since the vision space learnt by CLIP is guided by language supervision which is of high-level semantic meaning, the visual feature is rich in semantic information. Therefore, we use a pretrained CLIP model to extract the features of visual prompts.

\subsubsection{CLAP.}
CLAP~\cite{clap} is currently the largest audio representation FM trained with 2.5M text-audio paired data. Similar to CLIP, CLAP learns a joint text-audio embedding space via contrastive learning under the language supervision. A critical reason we choose CLIP and CLAP is that they both share the text modality as a common domain during their training. We assume text could serve as a bridge which makes the translation from vision to audio easier.

\subsubsection{AudioLDM.}
AudioLDM~\cite{audioldm} is a continuous latent diffusion model (LDM) trained in a self-supervised way with 3.3M 10-second audio clips. Conditioned on CLAP audio embedding, it generates the latent code of audio mel-spectrogram which can be decoded and converted into audio waveform. The original work only explores the usage of the LDM part in text-to-audio (T2A) generation task. Since CLAP represents text and audio jointly, AudioLDM can directly take text as input when being adapted to T2A task.  We note that despite being proposed specifically for T2A generation task, AudioLDM is expected to adapt more naturally to audio features. This inspires us to ponder if we could translate a vision feature into its corresponding audio embedding in CLAP space, then we could keep AudioLDM completely intact and utilize it as an off-the-shelf audio generator FM.

\begin{figure}[!t]
\centering     
\subfigure[]{\includegraphics[width=0.4\textwidth]{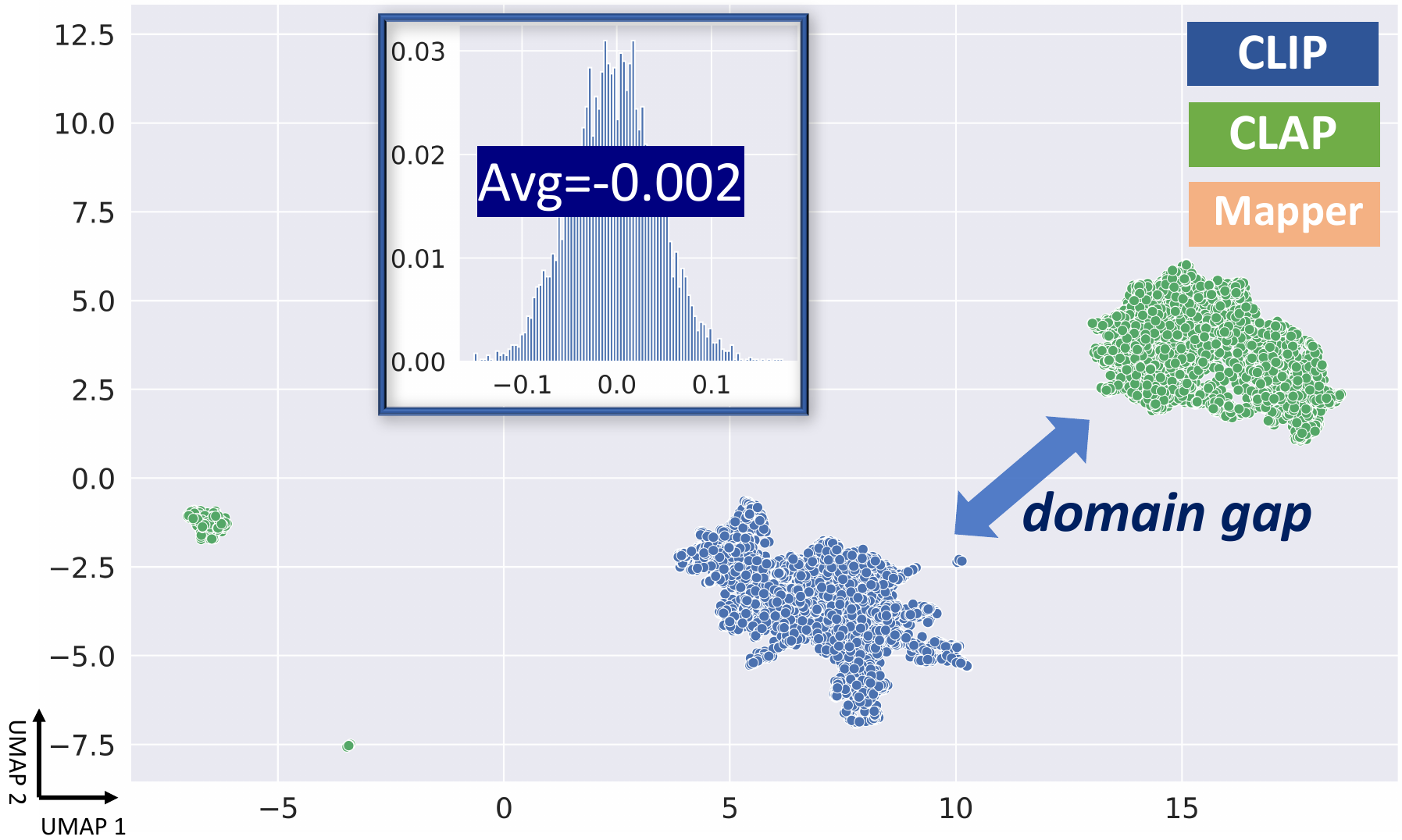}\label{fig:bridgea}}
\subfigure[]{\includegraphics[width=0.48\textwidth]{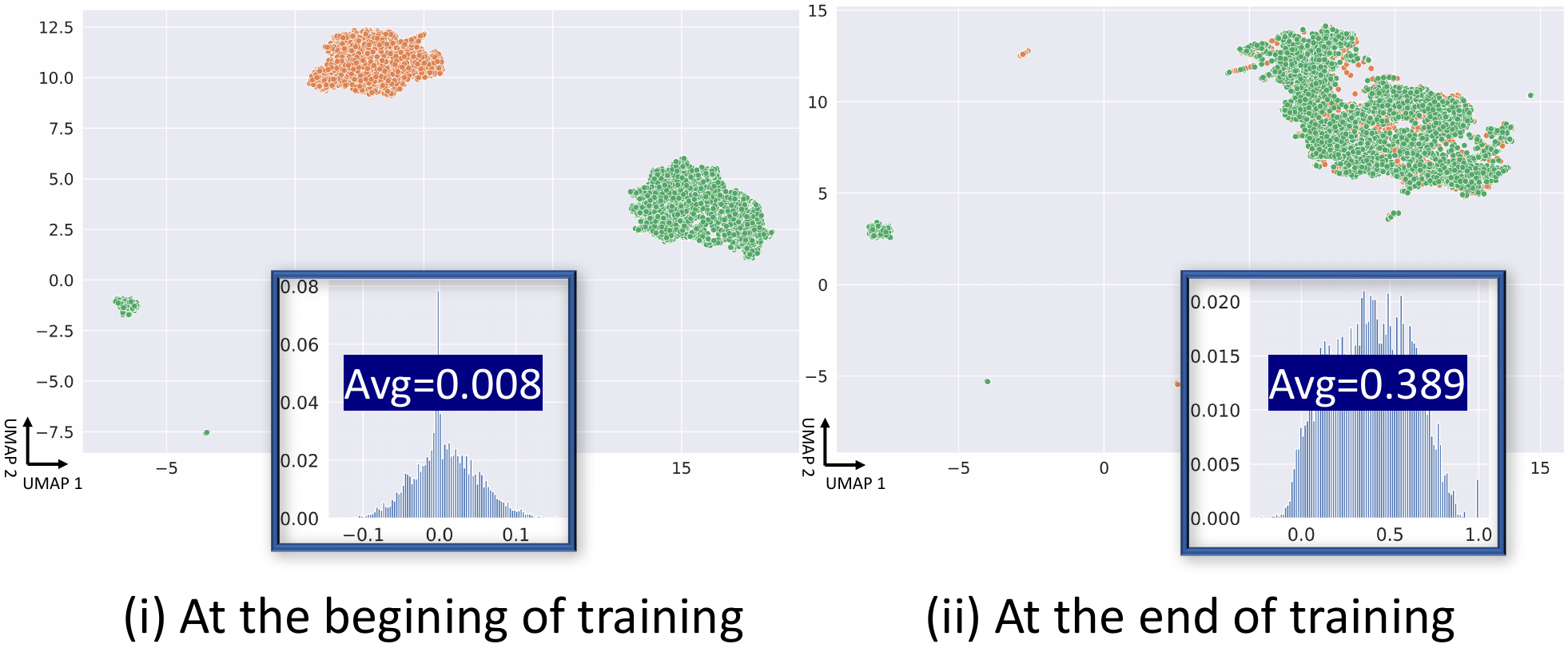}\label{fig:bridgeb}}
\caption{We visualize the domain gap between CLIP image space and CLAP audio space in (a). In (b), we present the process of closing the domain gap during the training of the V2A-Mapper. The accompanying histograms display the cosine similarity between paired embeddings from two domains. The larger the value is, the closer two domains are.}
\label{fig:bridge}
\end{figure}

\subsection{Bridge the Domain Gap between Vision and Audio}
\label{sec:bridge}
We first investigate if there exists a domain gap between the vision and audio spaces learnt by CLIP and CLAP respectively. Following ~\cite{mindthegap}, we measure it by randomly selecting 5000 samples from the video dataset VGGSound~\cite{vggsound} to get an estimation of both the visual and auditory feature distributions. Specifically, we encode the video frames into 512-d feature vectors with pretrained CLIP image encoder and average them along the time axis to get one single embedding for each video. For audio data, we project each audio sample into 512-d feature vector with pretrained CLAP audio encoder. We then use UMAP visualization~\cite{umap} to project 5000 CLIP image embeddings and 5000 CLAP audio embeddings into the same 2-d space. As shown in Fig.~\ref{fig:bridgea}, the average cosine similarity between paired visual (CLIP) and audio (CLAP) features is near 0 and there indeed exists a considerable gap between CLIP image domain and CLAP audio domain. 

To bridge the domain gap, we propose to train a mapper, namely V2A-Mapper, between CLIP and CLAP so that the visual embedding could be translated into the CLAP space. The upper part of Fig.~\ref{fig:overview} shows the training pipeline. A video $V_i$ is a sequence of $n$ images $\left\{V_i^1, ..., V_i^n\right\}$. To get the visual embedding for a video, we use frozen CLIP model to encode each frame into 512-d feature vector to get a set of frame features $\left\{(E^1_i)^v, ..., (E^n_i)^v\right\}$. We then use an aggregator function $\sigma$ to get a single vector $E^v_i$ as the visual feature for the video input. The aggregator function could be: 1) randomly picking one vector; 2) picking the vector of the middle frame; 3) averaging along time axis. According to the experiments, the third option obtains the best performance in both fidelity and relevance. Similarly, for the paired audio data $A_i$, we encode it into 512-d feature vector $E^a_i$ with frozen CLAP model. Once we have paired visual features $E^v_i$ and auditory features $E^a_i$, we can train the mapper to convert the CLIP embedding $E_i^v$ into a pseudo CLAP embedding $E_i^{a'}$. We use Mean Square Error loss to guide the training. The training process can be formulated as below:
\begin{align}
    L = \mathbb{E}_{i\sim\left[1,K\right]}\left[\parallel E^a_i - E_i^{a'}\parallel^2\right],
\label{mseloss}
\end{align}
where $K$ is the batch size and $E_i^{a'}$ is from $mapper(E^v_i)$. Fig.~\ref{fig:bridgeb} visualizes the domain shift after the training. Since the mapper is randomly initialized at the beginning, the translated embedding cluster is still far from the target CLAP space as displayed in Fig.~\ref{fig:bridgeb}(i). When training finishes, the translated space and the target CLAP space become overlapped as suggested in Fig.~\ref{fig:bridgeb}(ii) indicating the mapper is optimized successfully. 

     


\begin{table*}[!t]
\footnotesize
\centering
    \resizebox{0.71\linewidth}{!}{
    \begin{tabular}{l|lll|c|l|l}
    \toprule
    \multirow{2}[3]{*}{Method} & \multicolumn{3}{c|}{VGGSound} & ImageHear   & 
    \multirowcell{2}{Infer. Time (s) $\downarrow$} & \multirowcell{2}{\#Trainable Param. (M) $\downarrow$}\\
    \cmidrule(lr){2-4} \cmidrule(lr){5-5} 
          & FD $\downarrow$ & FAD $\downarrow$  & \multicolumn{1}{c|}{CS $\uparrow$}  & CS $\uparrow$   &  &  \\
         
    \midrule
    Reference   & 0 & 0 & 8.925	& - &  - &	- \\
    \midrule
    Im2Wav   & 51.500 & 6.005  & 7.827	& 9.843 &  864.12 &	360.40 \\
    CLIPSonic-IQ   & 27.124 & 3.495 & 7.251 & 11.392 &   53.94 & 142.58 \\
    Ours  & \textbf{24.168} & \textbf{0.841} & \textbf{9.720} &  \textbf{11.950} &  \textbf{35.45} &	\textbf{48.83}\\
    
    \bottomrule
    \end{tabular}
    }
    \resizebox{0.28\linewidth}{!}{
    \begin{tabular}{c}
    \includegraphics[width=1\linewidth]{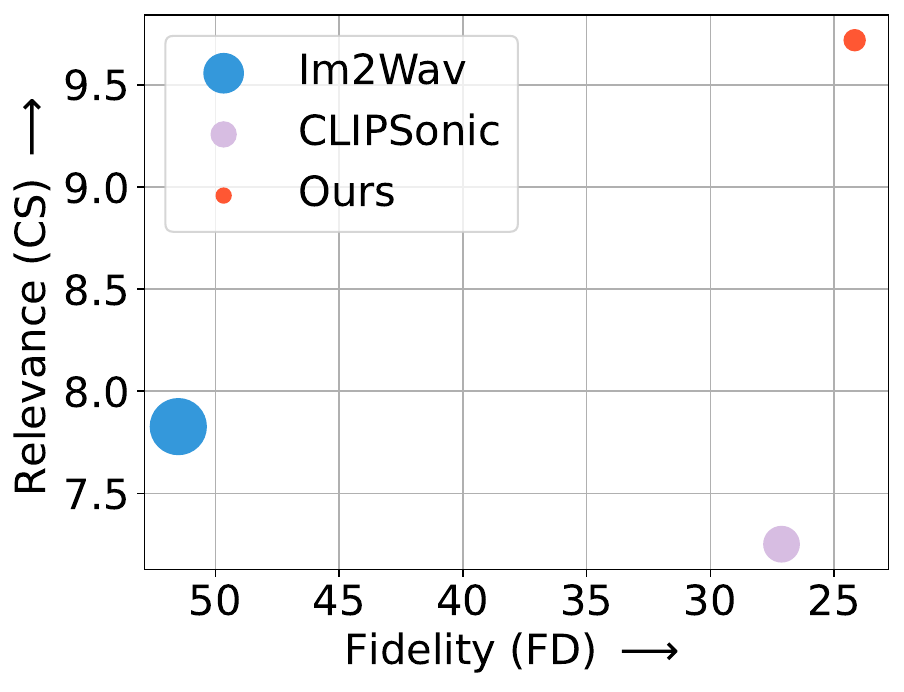}
    \end{tabular}}
    \vspace{-3mm}
    
    \caption{Objective comparison with SOTA methods on VGGSound (video-to-sound generation) and ImageHear (image-to-sound generation). The inference time is measured as the average time spent for 100 samples through the whole pipeline from input visual prompts to output waveforms on one NVIDIA RTX A6000 GPU. Our method achieves the best on all the objective metrics. We also plot the comparison on VGGSound in the right diagram to showcase our method achieves better results on both metrics and contains fewer trainable parameters (smaller circle size).} 
\label{tab:sota}
\end{table*}

\begin{table*}[!t]
\centering
\begin{tabular}{l|ll|ll}
\toprule
\multirow{2}[3]{*}{Method} & \multicolumn{2}{c|}{VGGSound} & \multicolumn{2}{c}{ImageHear}  \\
\cmidrule(lr){2-3} \cmidrule(lr){4-5} 
      & Fidelity $\uparrow$ & \multicolumn{1}{c|}{Relevance $\uparrow$}  & Fidelity $\uparrow$ & \multicolumn{1}{c}{Relevance $\uparrow$}  \\
     
\midrule
Reference   & 3.580$\pm$0.455 & 4.178$\pm$0.533	& - & - \\
\midrule
Im2Wav   & 1.838$\pm$0.511  & 2.415$\pm$0.645	& 1.840$\pm$0.502 & 2.705$\pm$0.398  \\
CLIPSonic-IQ   & 2.533$\pm$0.522  & 2.140$\pm$0.551 & 2.888$\pm$0.502 & 3.215$\pm$0.291 \\
Ours  & \textbf{2.845}$\pm$\textbf{0.491} & \textbf{2.808}$\pm$\textbf{0.651} &  \textbf{3.425}$\pm$\textbf{0.459} & \textbf{3.310}$\pm$\textbf{0.295} \\

\bottomrule
\end{tabular}
\caption{Subjective comparison with SOTA methods on VGGSound (video-to-sound generation) and ImageHear (image-to-sound generation). Our lightweight solution outperforms previous methods in both sound quality and the relevance to visual prompt from a human perception perspective.}
\label{tab:sota-mos}
\end{table*}

\subsection{Diffusion-based V2A-Mapper}
\label{sec:mapper}
Since the mapper is expected to project the embedding from visual space to audio space, a natural way to implement the mapper is a stack of multilayer perceptrons (MLPs) as a one-to-one regression task. Inspired by DALLE2's prior model~\cite{dalle2}, we consider the projection process as a conditional generation task, which models a one-to-many mapping ensuring the diversity and generalization of the target audio distribution. Specifically, we train the mapper as a diffusion model~\cite{dm_1,dm_1_1}. It includes a forward process where Gaussian noises are gradually added to the target audio embedding $E^a_{i,0}$ until it approaches to a standard Gaussian distribution $E^a_{i,T}$ (i.e., completely random) for $T$ timesteps and a reverse process where the target is gradually recovered from the noisy distribution by canceling the added noises with a network in a recursive manner. Following DALLE2, instead of predicting the intermediate noises added at each step~\cite{dm_1}, we directly predict the target audio embedding. Therefore, we train the mapper network $f_\theta$ to predict audio embedding $E^a_{i,0}$ based on the timestep $t$, the noisy audio embedding $E^a_{i,t}$ at timestep $t$, and the condition visual embedding $E_i^v$. Hence, the training objective in Eq.~\ref{mseloss} can be formulated as:
\begin{align}
    L = \mathbb{E}_{i\sim\left[1,K\right], t\sim\left[1, T\right]} \left[ \parallel E^a_{i,0}-f_\theta\left(t, E^a_{i,t},E_i^v\right)\parallel^2\right].
\label{diffusion}
\end{align}

We experiment with two different architectures for the mapper network $f_\theta$ - simple MLPs and Transformer. For the Transformer variant, we craft a learnable token of 512-d whose output from the Transformer is considered as the recovered audio embedding. We then take the time embedding, noisy audio embedding, as well as the visual condition as three other tokens of the same shape (i.e., 512-d) to the Transformer encoder to obtain the recovered audio embedding. For the simple MLP variant, we concatenate all the three tokens as input and output the final 512-d vector as the predicted audio embedding via fully-connected network. We find the Transformer is a better way to incorporate the condition compared to simple concatenation in MLPs.

\section{Experiments}
\subsection{Experimental Setup}
\subsubsection{Datasets.}
We train our V2A-Mapper and all the variants on VGGSound video dataset~\cite{vggsound}. VGGSound contains 199,176 10-second video clips extracted from
videos uploaded to YouTube with audio-visual correspondence. Note that VGGSound has never been used as training data for the foundation models we adapt. Following the
original train/test splits, we train on 183,730 videos and evaluate on 15,446 videos. To testify the generalization ability of our V2A-Mapper, we also test on out-of-distribution dataset ImageHear~\cite{im2wav} which contains 101 images from 30 visual classes (2-8 images per class). We generate 10-second audio samples for all the evaluations. 

\subsubsection{Metrics.}
We measure the performance on two aspects, fidelity and the relevance to the visual prompt. Specifically, we use Fréchet Distance (FD) to measure the overall quality and variability of generated audio clips. FD computes the distance of embedding distributions between the synthesized and the real samples. To compare with previous methods, we also compute the Fréchet Audio Distance (FAD)~\cite{fad}. FD and FAD differ at the embedding extractor - FD uses PANNs~\cite{panns} while FAD adopts VGGish~\cite{vggish}. Similar to ~\cite{audioldm}, we choose FD as our main evaluation metric regarding the sound quality since PANNs is superior to VGGish by considering long distance temporal change. For the relevance evaluation, we use CLIP-Score (CS)~\cite{im2wav} to get the cosine similarity between the CLIP embedding of the visual input and the Wav2CLIP~\cite{wav2clip} embedding of the generated sound. As Wav2CLIP learns an audio encoder via contrastive loss on VGGSound with the guidance of frozen CLIP image encoder, if the generated sound matches the visual input, the Wav2CLIP embedding is expected to be similar to its paired CLIP embedding. 

\subsubsection{Subjective Testing.}
To complement the objective metrics, we also conduct a listening test to measure the fidelity of the generated audio clips and their relevance to visual prompts from a human perception perspective. We ask 20 listeners to rate audio clips of 20 randomly selected visual samples on a discrete 5-point scale in terms of fidelity and relevance, respectively. The average rating across all listeners for each algorithm is computed as Mean Opinion Score (MOS)~\cite{mos_correct}. We also re-code the responses into paired comparisons and infer the relative standings via indirect scaling~\cite{hannes_rankorder}. We calculate the degree by which other approaches exceed our method as the Just Meaningful Difference (JMD) score (e.g. a negative value indicates inferiority of other algorithms compared to ours). More details of our human evaluation are provided in the supplementary.

\subsubsection{Implementation Details.} 
We use ``ViT-B/32" version for CLIP model\footnote{https://github.com/openai/CLIP}. For CLAP model and audio generator, we use pretrained models from AudioLDM\footnote{https://github.com/haoheliu/AudioLDM}. For the diffusion-based V2A-Mapper, we use a cosine noise schedule with 1000 diffusion steps during training and 200 steps at inference time. We use AdamW with a learning rate of 1.1e-4, a batch size of 448 visual-audio embedding pairs, and a dropout rate of 0.1 in classifier-free guidance. We provide more implementation details including datasets used in adopted FMs, the full experiments of architecture hyperparameter tuning, and the guidance scale tuning in the supplementary. 

\begin{figure}[!t]
\centering     
\subfigure[Fidelity.]{\includegraphics[width=0.225\textwidth]{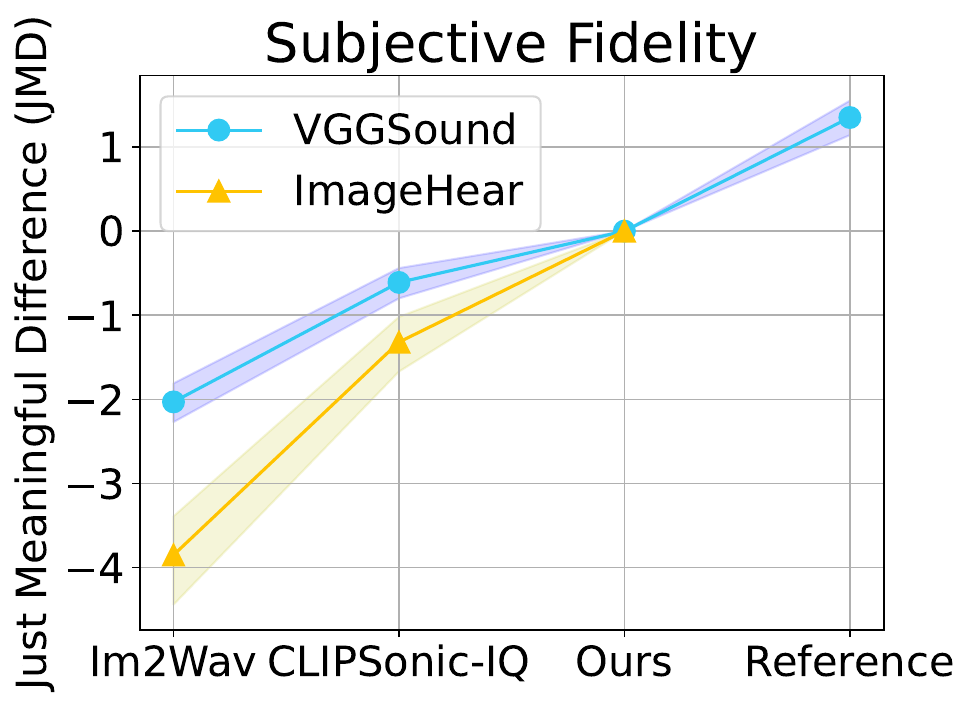}\label{fig:st_im2sd}}
\subfigure[Relevance.]{\includegraphics[width=0.225\textwidth]{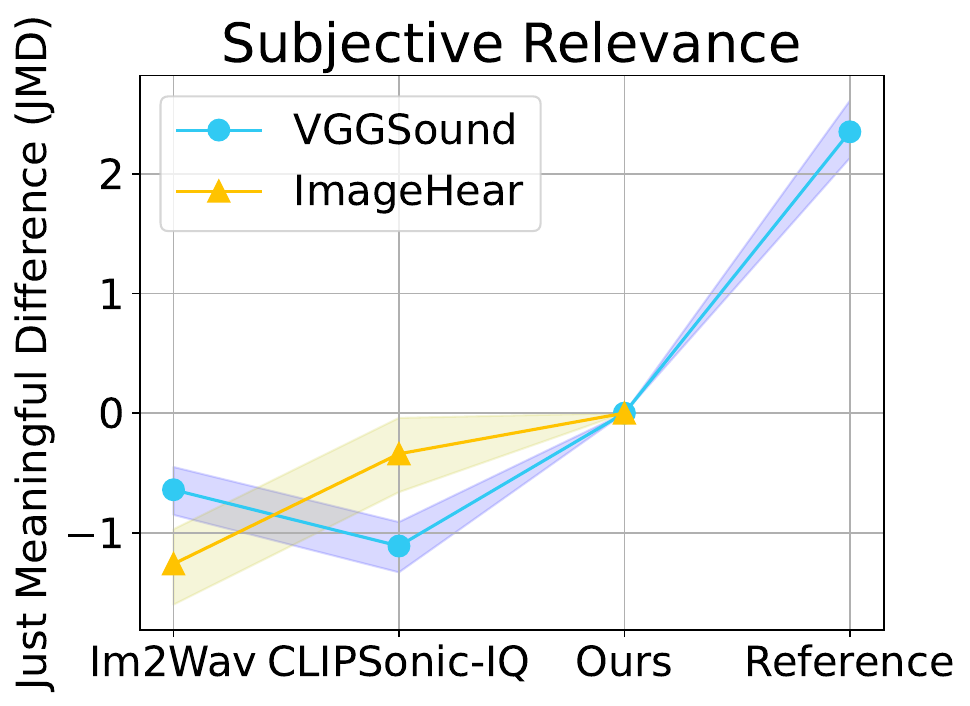}\label{fig:st_vd2sd}}
\caption{The Just Meaningful Difference steps of current methods relative to our algorithm with 95\% bootstrap confidence intervals.}
\label{fig:subj_is}
\end{figure}

\subsection{Compare with SOTA}
Im2Wav~\cite{im2wav} is the current state-of-the-art method in open-domain vision-to-audio generation. It involves training of two Transformer decoders of different scales for latent code generation, a VQ-VAE for audio mel-spectrogram encoding and decoding, and a vocoder for waveform conversion. CLIPSonic-IQ~\cite{clipsonic} is a concurrent work to ours and they train a diffusion model to directly generate mel-spectrogram conditioned on visual representation. They also require the training of a BigVGAN~\cite{bigvgan} to convert the generated mel-spectrogram into audio waveform. Compared to these methods, our approach only requires the training of a single V2A-Mapper. Trained with the same modestly sized VGGSound data, our method achieves better performance as a result of the knowledge transfer from foundation models. 

\subsubsection{Objective Results.}
Tab.~\ref{tab:sota} shows that the proposed method achieves superior performance in all objective metrics. Compared to Im2Wav, our method trains with 86\% fewer parameters but achieves 53\% and 19\% improvement in FD and CS, respectively. It is also noticeable that our method is significantly faster than Im2Wav (x24 faster) during inference. Our method also outperforms CLIPSonic-IQ in all the metrics with fewer parameters and faster inference speed. Note that our method exceeds even the reference for the relevance metric (CS). We conjecture that this is because VGGSound contains noisy data whose audio and visual streams might not be highly-relevant, which could suggest the proposed method is robust to noisy training data. We recommend readers to watch the sports live video in our demo website to observe this phenomenon. 

\subsubsection{Subjective results.}
As shown in Tab.~\ref{tab:sota-mos} and Fig.~\ref{fig:subj_is}, our method exceeds previous works in both fidelity and relevance. We notice that the usage of diffusion model could especially boost the audio quality as indicated by the improvement achieved by both CLIPSonic-IQ and our method. While CLIPSonic-IQ fails at the relevance aspect when taking videos as input, our method consistently outperforms the SOTA method on both videos and images. However, we note that there is still a gap between our performance and the ground truth. Empirically, we find temporal alignment to be a major issue that leads to unsatisfactory relevance rating, which we will attempt to address in our future work. 

\begin{figure}[!t]
\centering     
\subfigure[vision-text-audio.]{\includegraphics[width=0.5\textwidth]{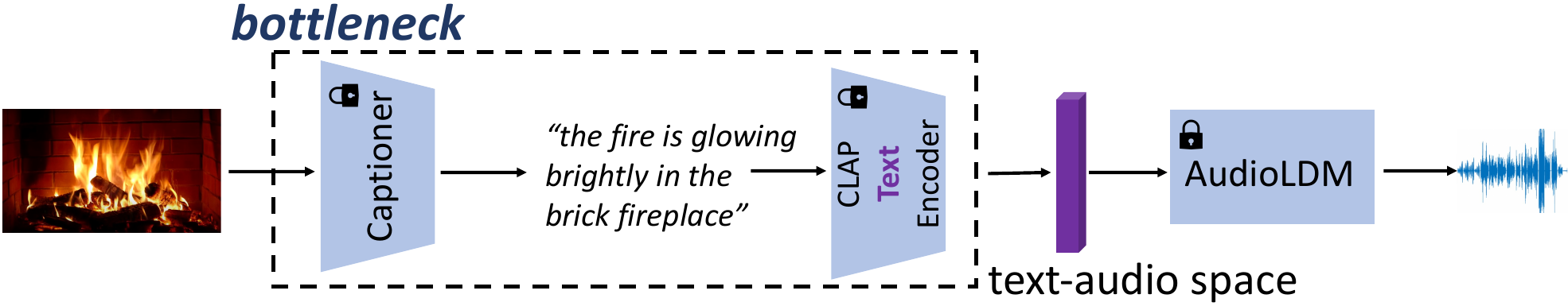}\label{fig:fm_w_captioner}}
\subfigure[w/o mapper.]{\includegraphics[width=0.4\textwidth]{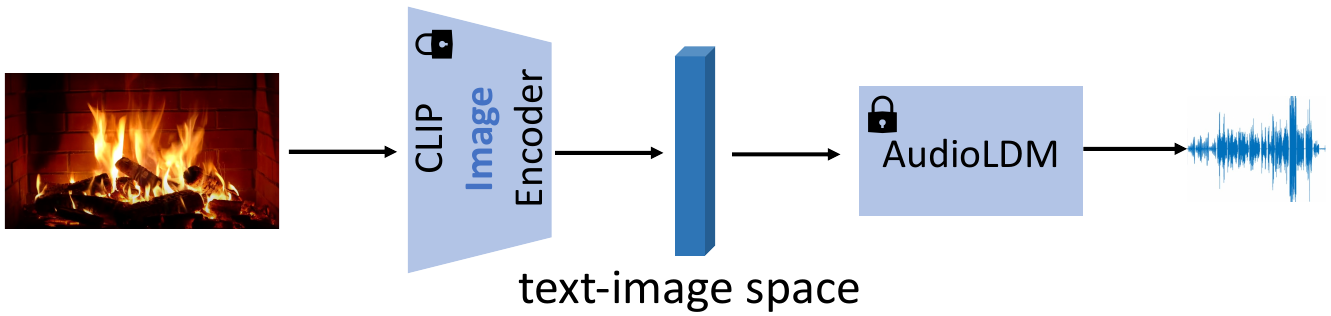}\label{fig:fm_wo_mapper}}
\caption{Different ways of using FMs. (a) adopts a captioner to generate text description as a bridge to use AudioLDM while (b) directly puts the visual features from CLIP image encoder as the condition to AudioLDM.}
\label{fig:abl-fms}
\end{figure}

\begin{table}[!t]
\centering
\begin{tabular}{l|ll|c}
\toprule
\multirow{2}[3]{*}{Method} & \multicolumn{2}{c|}{VGGSound} & ImageHear   \\
\cmidrule(lr){2-3} \cmidrule(lr){4-4} 
      & FD $\downarrow$ & \multicolumn{1}{l|}{CS $\uparrow$}  & CS $\uparrow$   \\
     
\midrule
vision-txt-audio   & 56.397  & 6.672	& 7.310  \\
w/o mapper   & 72.527  & 5.258 & 4.026 \\
w/ mapper  & \textbf{24.168} & \textbf{9.720} &  \textbf{11.950}\\

\bottomrule
\end{tabular}
\caption{Ablation study with different ways of using FMs for vision-to-audio synthesis.}
\label{tab:abl-fms}
\end{table}

\subsection{Ablation Study}
\subsubsection{Different Ways of Utilizing FMs.}
Since AudioLDM is proposed for text-to-audio generation, the naive way of utilizing it for V2A synthesis is interleaving a captioning model to generate text input as shown in Fig.~\ref{fig:fm_w_captioner}. To verify this vision-txt-audio idea, we adopt SOTA captioner BLIP~\cite{blip} to generate descriptions for images. For video-to-audio generation, we use the tag information provided in VGGSound. As reported in Tab.~\ref{tab:abl-fms}, although using text as bridge could mitigate the gap to some extent, it is still inferior to our method with the V2A-Mapper in both fidelity and relevance. This result indicates that the captioner is actually a bottleneck whose performance would directly affect what audio is to be generated from AudioLDM. We provide two examples where the captioner fails at predicting correct object category in ``Why Vision-Text-Audio is Bottlenecked by the Captioner" section of our demo website. We suggest readers to check the audio results to examine the bottleneck challenge. Instead of decoding the visual condition into text format, our V2A-Mapper keeps the visual information as its latent code form and explicitly translates it from CLIP's visual space to CLAP's audio space, which could avoid information loss occurred during vision-txt-audio conversion. If the V2A-Mapper is skipped as illustrated in Fig.~\ref{fig:fm_wo_mapper}, the domain gap between vision and audio space prevents AudioLDM from generating high-fidelity and visually-relevant sound. Audio examples showcasing the difference are presented in ``Domain Gap Bridging Process" section of our demo website. The recent text-to-audio generation work Make-An-Audio~\cite{makeanaudio} trained their audio generator with CLIP text embedding and adopted CLIP image embedding as input to handle vision-to-audio generation. Similar to the ``w/o mapper" strategy, the domain gap between the visual condition and the target embedding space which their audio generator works on is not addressed. We refer readers to our demo website to observe the comparison with Make-An-Audio.

\subsubsection{Inside the Mapper: Generative vs. Regression.}
The V2A-Mapper can be implemented in a generative or a regression strategy. A generative V2A-Mapper learns a one-to-many mapping while a regression one builds a one-to-one projection. As displayed in Tab.~\ref{tab:abl-mapper}, although regression model could learn a slightly better relevance due to the one-to-one mapping, the generated sound lacks diversity and fidelity as suggested by much worse FD scores. A generative mapper is critical to ensure the variability as also observed in text-to-image synthesis~\cite{dalle2}. To showcase the diversity of our method, we provide three samples for each visual input in the ``Variability of Our V2A Generation Model" section of our demo website. And compared to linear projections, the attention mechanism used in Transformer could integrate the visual condition in a better way. 

\begin{table}[!t]
\small
\setlength\tabcolsep{3pt}
\centering
\begin{tabular}{l|l|ll|c}
\toprule
\multicolumn{2}{c|}{\multirow{2}[3]{*}{Arch. of the V2A-Mapper}} & \multicolumn{2}{c|}{VGGSound} & ImageHear   \\
\cmidrule(lr){3-4} \cmidrule(lr){5-5} 
 \multicolumn{2}{c|}{} & FD $\downarrow$ & \multicolumn{1}{l|}{CS $\uparrow$}  & CS $\uparrow$   \\
     
\midrule
\multirow{2}{*}{Regression} & MLPs  & 35.059  & 9.927	& 12.048  \\
& Transformer   & \textbf{29.378}  & \textbf{10.076} & \textbf{12.317} \\
\midrule
\multirow{2}{*}{Generative} & diff. w/ MLPs  & 28.803 & 8.685 &  10.449\\
& diff. w/ Transformer & \textbf{24.168} & \textbf{9.720} &  \textbf{11.950}\\

\bottomrule
\end{tabular}
\caption{Ablation study with different V2A-Mapper strategies (regression vs. generative) and architectures (MLPs vs. Transformer).}
\label{tab:abl-mapper}
\end{table}

\begin{table}[!t]
\centering
\begin{tabular}{l|ll|c}
\toprule
\multirow{2}[3]{*}{Aggregation} & \multicolumn{2}{c|}{VGGSound} & ImageHear   \\
\cmidrule(lr){2-3} \cmidrule(lr){4-4} 
      & FD $\downarrow$ & \multicolumn{1}{l|}{CS $\uparrow$}  & CS $\uparrow$   \\
     
\midrule
random   & 24.826  & 9.200	& 11.465  \\
middle   & 25.569  & 9.192 & 11.901 \\
average  & \textbf{24.168} & \textbf{9.720} &  \textbf{11.950}\\

\bottomrule
\end{tabular}
\caption{Ablation study with different ways of aggregation for video feature representation during training.}
\label{tab:abl-aggregation}
\end{table}

\subsubsection{Different Aggregator Methods $\sigma$.}
We explore three different ways of aggregating visual information of videos: 1) randomly select one frame as the key frame to represent the video; 2) instead of using random frame, choose the middle one; 3) average the CLIP features of all the frames along the time axis. Tab.~\ref{tab:abl-aggregation} shows the performance of models trained with different aggregation methods. Since the task is to generate a large time-span (10 seconds) of a highly dynamic signal (audio), having time-related information in the condition could help. The average of abstract frame embeddings with rich semantic contents throughout the temporal dynamics is a better summary of the video than a single frame.

\subsubsection{Different Pretrained Vision FMs.}
\begin{table}[!t]
\small
\setlength\tabcolsep{3pt}
\centering
\begin{tabular}{l|l|ll|c}
\toprule
\multicolumn{2}{c|}{\multirow{2}[3]{*}{Method}} & \multicolumn{2}{c|}{VGGSound} & ImageHear   \\
\cmidrule(lr){3-4} \cmidrule(lr){5-5} 
 \multicolumn{2}{c|}{} & FD $\downarrow$ & \multicolumn{1}{l|}{CS $\uparrow$}  & CS $\uparrow$   \\
     
\midrule
\multirow{2}{*}{w/o mapper} & BLIP  & \textbf{53.621}  & 4.948	& \textbf{4.314}  \\
& CLIP   & 72.527  & \textbf{5.258} & 4.026 \\
\midrule
\multirow{2}{*}{w/ mapper} & BLIP  & 24.788 & 9.402 &  10.836\\
& CLIP & \textbf{24.168} & \textbf{9.720} &  \textbf{11.950}\\

\bottomrule
\end{tabular}
\caption{Ablation study with different vision-language models.}
\label{tab:abl-vl}
\end{table}
Our V2A-Mapper can be generalized to other vision-language models such as BLIP~\cite{blip}. As shown in Tab.~\ref{tab:abl-vl}, the proposed V2A-Mapper can boost the performance of both CLIP- and BLIP-based systems. We also note that no matter what vision-language model is used and how big the domain gap between the vision and audio spaces is, the proposed V2A-Mapper can bridge the gap and translate visual information into audio space - two systems achieve similar performance with the proposed V2A-Mapper. 

\subsubsection{Different Pretrained Audio FMs.}
\begin{table}[!t]
\centering
\small
\setlength\tabcolsep{1pt}
\begin{tabular}{l|ll|c|l|l}
\toprule
\multirow{2}[3]{*}{Method} & \multicolumn{2}{c|}{VGGSound} & ImageHear   & 
\multirowcell{2}{Time (s) $\downarrow$} & \multirowcell{2}{\#Param. (M) $\downarrow$}\\
\cmidrule(lr){2-3} \cmidrule(lr){4-4} 
      & FD $\downarrow$ & \multicolumn{1}{c|}{CS $\uparrow$}  & CS $\uparrow$   &  &  \\
     
\midrule
audioldm-s   & 25.635  & 9.547	& 11.586 &  9.33 &	185.04 \\
audioldm-s-v2 & \textbf{24.168} & \textbf{9.720} & 11.950 &  \textbf{9.33} &	\textbf{185.04}\\  
audioldm-l  & 25.130  & 9.531 & \textbf{12.016} &   11.58 & 739.14 \\

\bottomrule
\end{tabular}
\caption{Ablation study with different pretrained AudioLDM models.}
\label{tab:abl-audiofm}
\end{table}
We ablate with different pretrained audio generators from AudioLDM: 1) audioldm-s is the base model; 2) audioldm-s-v2 is the base model but trained with more steps; 3) audioldm-l is the model with larger architecture. As shown in Tab.~\ref{tab:abl-audiofm}, either scaling the model up or optimizing
its training for longer steps can help enhance the performance to some extent. Therefore, we hypothesize that a better audio generator FM could further improve the quality and relevance in the future. 

\subsection{Latent Space Interpolation}
\begin{figure}[!t]
\centering     
\subfigure[Guided by image.]{\includegraphics[width=0.2\textwidth]{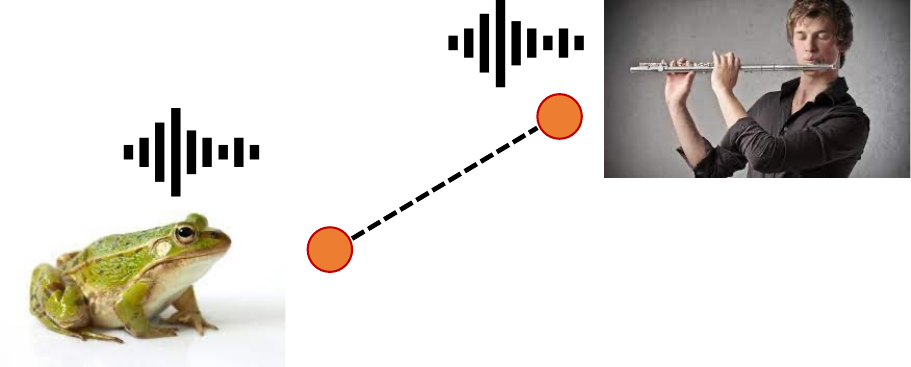}\label{fig:frog-flute}}
\subfigure[Guided by text.]{\includegraphics[width=0.23\textwidth]{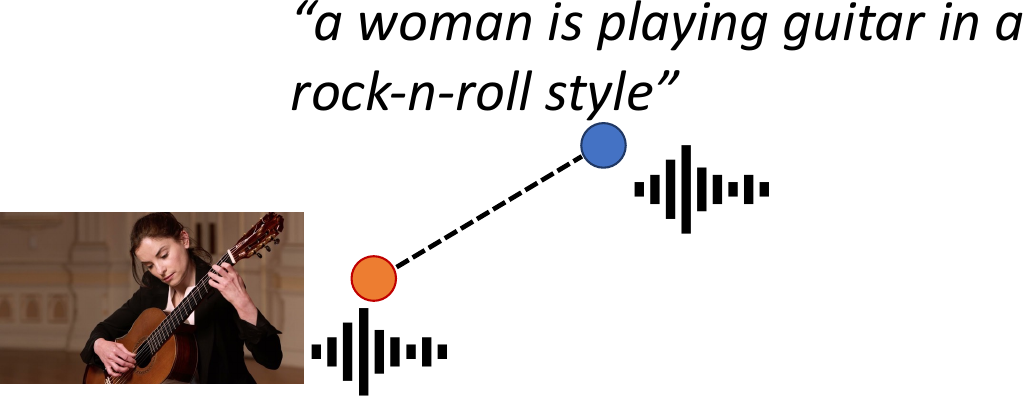}\label{fig:guitar-txt}}
\caption{Our V2A-Mapper enables interpolation guided by both image and text. Audios are provided in demo website.}
\label{fig:interpolation}
\end{figure}

As the visual condition is translated into the CLAP latent space, we could interpolate audio embeddings by either visual or textual guidance. For simplicity, we perform linear interpolation between two embeddings. As shown in  Fig.~\ref{fig:interpolation}, the interpolation can happen from a frog sound to a sound indicated by an image of a man playing flute, or to a target specified by a description. It is noticeable that vision, text, and audio are semantically gathered to the same space without actual training with three modalities. We hear a relatively smooth transition during the interpolation, which indicates auditorily that the V2A-Mapper does learn the translation from CLIP space to CLAP space. Examples are provided in ``Latent Space Interpolation" section of our demo website.

\section{Discussion}
\subsubsection{Limitation and Future Work.}
While our approach has achieved considerable success, it is important to acknowledge several limitations. First, the system can not achieve finer control. The generated sound exhibits semantic relevance in a general sense, but it lacks controllability over specific details. Second, the system fails when the visual cues involve unclear subjects (e.g., multiple objects, blurry/damaged images). Third, the system does not explicitly handle the temporal alignment between audio and visual signals. All of these could be interesting future directions in this area. Enforcing text into the condition could be a starting point for explicit controllability by considering both visual and textual features. To incorporate the language information of high-level semantic meaning into the system, recent multimodal foundation models such as Meta-Transformer~\cite{meta_transformer} and VATLM~\cite{zhu2023vatlm} could be taken into consideration. They learn the representation across vision, language, and audio which could shape a common space for different modalities. 
\subsection{Ethical Statement.}
Our method aims to leverage foundation models for efficient vision-to-audio generation. It can be used to enhance the immersion of human experience, such as video editing and foley design. Nevertheless, the application of this technology poses a risk if being maliciously misused on social platforms, potentially resulting in negative outcomes for society. Although significant strides have been made in audio deepfake detection research to mitigate such concerns~\cite{yamagishi21_asvspoof}, the availability of ample datasets remains pivotal for improving detection accuracy. In light of this, we are committed to presenting our synthesized audio samples, intending to contribute to the advancement and fine-tuning of existing detection algorithms.

\section{Conclusion}
In this paper, we explore the feasibility and efficiency of adapting foundation models (FMs) in the challenging open-domain vision-to-audio generation task. We propose a simple yet effective mapper mechanism (V2A-Mapper) to connect the representative visual FM CLIP and the generative auditory FM AudioLDM. Learning to translate visual features from CLIP space to the auditory CLAP space, the V2A-Mapper successfully passes visual information to its auditory counterpart from which the AudioLDM can synthesize high-fidelity and visually-aligned sound. Our method is relatively lightweight to train because it only requires optimization of the V2A-Mapper. Despite this simplicity, it achieves superior performance compared to current state-of-the-art approaches with far more complex training regimes as demonstrated by both subjective and objective evaluation.

\section{Acknowledgments}
We extend our sincere appreciation to Xiaoyu Liu, Hannes Muesch, Adam Mater, Chunghsin Yeh, Michael Eckert, and Lie Lu for their valuable comments, corrections and inspiration. In addition, we thank Xiaoyu for providing the code of CLIPSonic-IQ and we are grateful to Hannes and Adam for discussing, designing, and analyzing the subjective testings.

\bibliography{aaai24}

\clearpage
\begin{appendices}
In this supplementary, we provide more details of our experiments including implementation setup and hyper-parameter tuning. We also elaborate more on how we conduct the subjective testing in this document.
\section{Implementation Details}
All experiments are performed on a platform of four NVIDIA RTX A6000 GPUs. We train all the models with 100 epochs. The best model (diffusion w/ transformer) takes around 14 hours to train on four GPUs. Depending on architectures and strategies, training of other ablated models take from 40 minutes to 12 hours with one or four GPUs. We implement the network of V2A-Mapper in either Transformer or simple MLP. The transformer is implemented as multiple stacks of attention and feedforward layers. The number and dimension of attention heads is 12 and 64, respectively. The expansion rate for the feedforward network is set as 4. We ablate with different depths of the Transformer and empirically find 12 achieves the best result. For the simple MLP architecture, we implement it as a stack of linear projection followed by SiLU and LayerNorm. We ablate with different depths $D$ and expansion rate $E$ of intermediate layers. For the timestep embedding in diffusion models, we use learnable embedding followed by a linear projection layer to encode the scalar timestep information. Code will be released for benchmarking. For the training set of pretrained foundation models, CLIP is trained on 400M image-text pairs scrambled from internet~\cite{clip}. CLAP~\cite{clap} is trained on 2.5M text-audio pairs including LAION-Audio-630K\footnote{https://github.com/LAION-AI/audio-dataset/blob/main/laion-audio-630k/README.md}, AudioSet~\cite{audioset}, AudioCaps~\cite{audiocaps}, and Clotho~\cite{clotho}. AudioLDM~\cite{audioldm} is trained on 3.3M 10-second sound clips including AudioSet, AudioCaps, Freesound\footnote{https://freesound.org/}, and BBC Sound Effect\footnote{https://sound-effects.bbcrewind.co.uk/
search}. 



\section{Hyper-parameter Tuning}
In this section, we first present how we tune the architecture hyper-parameters to find the best model for each V2A-Mapper variant. We then show how the guidance scale of AudioLDM and our generative V2A-Mapper could influence the performance.

\begin{table}[!t]
\centering
\small
\begin{tabular}{l|l|llc|l}
\toprule
\multicolumn{2}{c|}{Reg. MLP}  & \multicolumn{2}{c|}{VGGSound} & ImageHear & \multirowcell{2}{\#Param. (M)}\\
\cmidrule(lr){1-2} \cmidrule(lr){3-4} \cmidrule(lr){5-5} 
 D & E     & FD $\downarrow$ & \multicolumn{1}{c|}{CS $\uparrow$}  & CS $\uparrow$   &   \\
     
\midrule
1 &  1  & 35.662  & 9.714	& 12.292 & 0.53 \\
1 & 2 & 35.336 & 9.724 & \textbf{12.718} & 1.05\\  
1 & 4 & \textbf{35.059} & \textbf{9.927} & 12.048 & 2.10\\  
2 & 1  & 36.934  & 9.065 & 11.604 & 0.79 \\
2 & 2 & 36.582 & 9.699 & 12.067 & 2.10\\  
2 & 4 & 36.041 & 9.732 & 11.600 & 6.30\\  
4 & 1  & 39.855  & 9.292 & 10.600 & 1.32 \\
4 & 2  & 41.524  & 9.299 & 10.296 & 4.21 \\
4 & 4  & 41.621  & 8.580 & 9.775 & 14.71 \\

\bottomrule
\end{tabular}
\caption{Hyper-parameter tuning of the depth (D) and expansion rate (E) of the regression-based MLP.}
\label{tab:reg_mlp}
\end{table}

\begin{table}[!t]
\centering
\small
\begin{tabular}{l|l|llc|l}
\toprule
\multicolumn{2}{c|}{Diff. MLP}  & \multicolumn{2}{c|}{VGGSound} & ImageHear & \multirowcell{2}{\#Param. (M)}\\
\cmidrule(lr){1-2} \cmidrule(lr){3-4} \cmidrule(lr){5-5} 
 D & E     & FD $\downarrow$ & \multicolumn{1}{c|}{CS $\uparrow$}  & CS $\uparrow$   &   \\
     
\midrule
1 &  1  & 30.814  & \textbf{8.720}	& 8.771 & 1.56 \\
1 & 2 & 30.023 & 8.655 & 9.465 & 2.61\\  
1 & 4 & 28.803 & 8.685 & \textbf{10.449} & 4.71\\  
2 & 1  & 29.456  & 8.675 & 9.484 & 1.83 \\
2 & 2 & 27.260 & 8.696 & 9.702 & 3.67\\  
2 & 4 & 26.582 & 8.553 & 10.027 & 8.91\\  
4 & 1  & 28.136  & 8.667 & 9.219 & 2.35 \\
4 & 2  & 27.461  & 8.368 & 9.760 & 5.77 \\
4 & 4  & \textbf{26.485}  & 8.605 & 9.839 & 17.32 \\

\bottomrule
\end{tabular}
\caption{Hyper-parameter tuning of the depth (D) and expansion rate (E) of the diffusion-based MLP.}
\label{tab:diff_mlp}
\end{table}
\subsection{Architectures}

\subsubsection{MLP.}
We experiment with both regression-based and diffusion-based V2A-Mapper using MLP as building block. Tab.~\ref{tab:reg_mlp} and Tab.~\ref{tab:diff_mlp} display the results of different depth and expansion rates for regression-based and diffusion-based strategies, respectively. Since we also take the timestep into consideration in diffusion-based model, the trainable parameters are slightly more than those in the regression-based model given the same hyper-parameter. For the regression-based MLPs, when depth is 1 and expansion rate is 4, they achieve the best results on VGGSound and the third best on ImageHear. Since the best performance for different metrics is achieved by different hyper-parameters, we report the results from the same combination as that of regression-based MLP in the main paper for simplicity.

\begin{table}[!b]
\centering
\small
\begin{tabular}{l|llc|l}
\toprule
Reg. Trans.  & \multicolumn{2}{c|}{VGGSound} & ImageHear & \multirowcell{2}{\#Param. (M)}\\
\cmidrule(lr){1-1} \cmidrule(lr){2-3} \cmidrule(lr){4-4} 
depth  & FD $\downarrow$ & \multicolumn{1}{c|}{CS $\uparrow$}  & CS $\uparrow$   &   \\
     
\midrule
 3  & 31.350  & 10.163 & 11.646 & 12.31 \\
6 & 30.206 & \textbf{10.227} & 11.864 & 24.31\\  
8  & \textbf{29.378}  & 10.076 & \textbf{12.317} &   32.31 \\
12  & 29.820  & 9.887 & 12.191 &   48.32 \\

\bottomrule
\end{tabular}
\caption{Hyper-parameter tuning of the number of depth for the regression-based Transformer.}
\label{tab:reg_trans}
\end{table}

\begin{table}[!t]
\centering
\small
\begin{tabular}{l|llc|l}
\toprule
Diff. Trans.  & \multicolumn{2}{c|}{VGGSound} & ImageHear & \multirowcell{2}{\#Param. (M)}\\
\cmidrule(lr){1-1} \cmidrule(lr){2-3} \cmidrule(lr){4-4} 
depth  & FD $\downarrow$ & \multicolumn{1}{c|}{CS $\uparrow$}  & CS $\uparrow$   &   \\
     
\midrule
 3  & 25.412  & 9.259 & 10.259 & 12.82 \\
6 & 25.869 & 9.546 & \textbf{12.062} & 24.82\\  
8  & 24.995  & 9.661 & 11.594 &   32.83 \\
12  & \textbf{24.168}  & \textbf{9.720} & 11.950 &   48.83 \\

\bottomrule
\end{tabular}
\caption{Hyper-parameter tuning of the number of depth for the diffusion-based Transformer.}
\label{tab:diff_trans}
\end{table}

\subsubsection{Transformer.}

Similar to experiments with MLPs, both regression-based and generative-based V2A-Mapper with Transformer as building block are experimented with. Tab.~\ref{tab:reg_trans} and Tab.~\ref{tab:diff_trans} show the influence of different numbers of layers to the performance for regression-based and diffusion-based V2A-Mapper, respectively. We set depth as 8 for the regression-based Transformer and depth as 12 for the diffusion-based Transformer.

\begin{figure}[!t]
\centering     
\subfigure[FD.]{\includegraphics[width=0.23\textwidth]{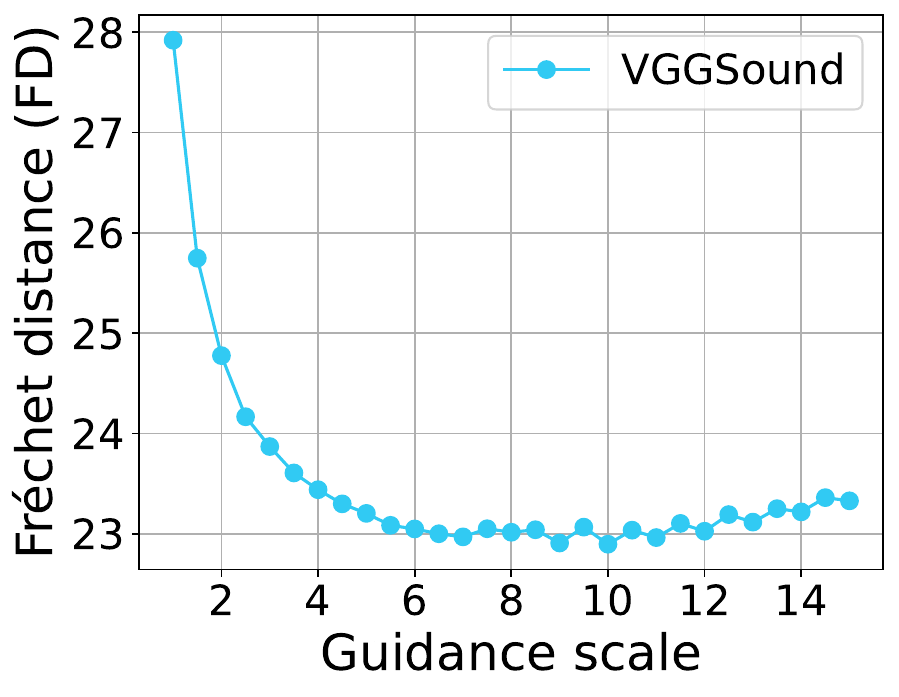}\label{fig:fd}}
\subfigure[CS.]{\includegraphics[width=0.23\textwidth]{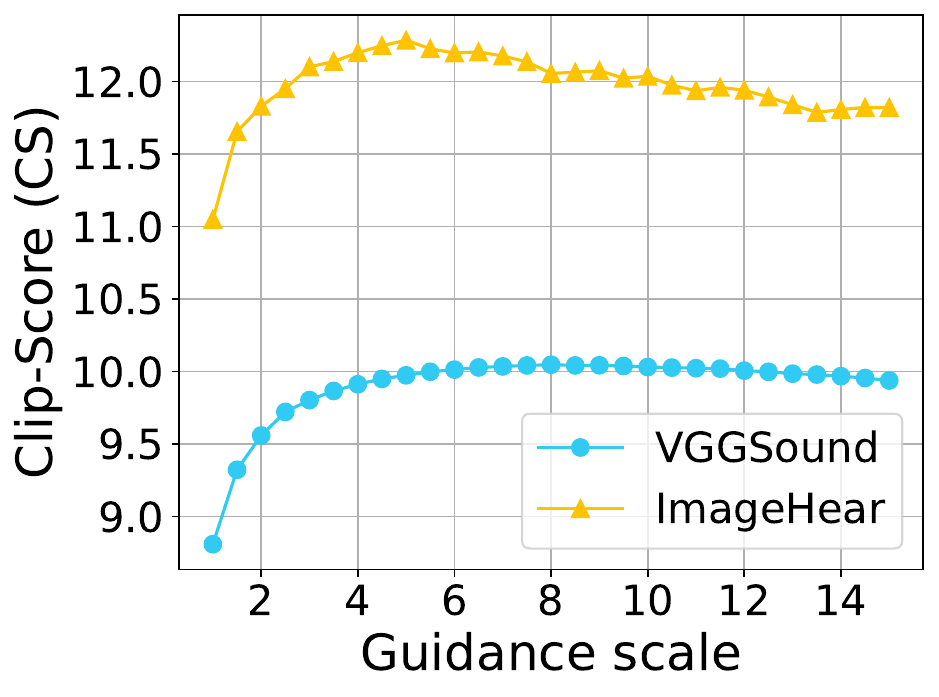}\label{fig:cs}}
\caption{Ablation study with different guidance scales of AudioLDM.}
\label{fig:gs}
\end{figure}
\subsection{Guidance Scale}
The guidance scale of a conditional diffusion model controls the degree of how much a condition should be taken into consideration~\cite{guidance1}. In this part, we investigate how the guidance scale of the audio generator AudioLDM and the generative V2A-Mapper affect the final performance. Tuning the guidance scale of both AudioLDM and our V2A-Mapper, the performance can be further boosted to 23.468 in FD for VGGSound and 9.967/12.370 in CS for VGGSound/ImageHear.

\begin{figure}[!t]
\centering     
\subfigure[FD.]{\includegraphics[width=0.23\textwidth]{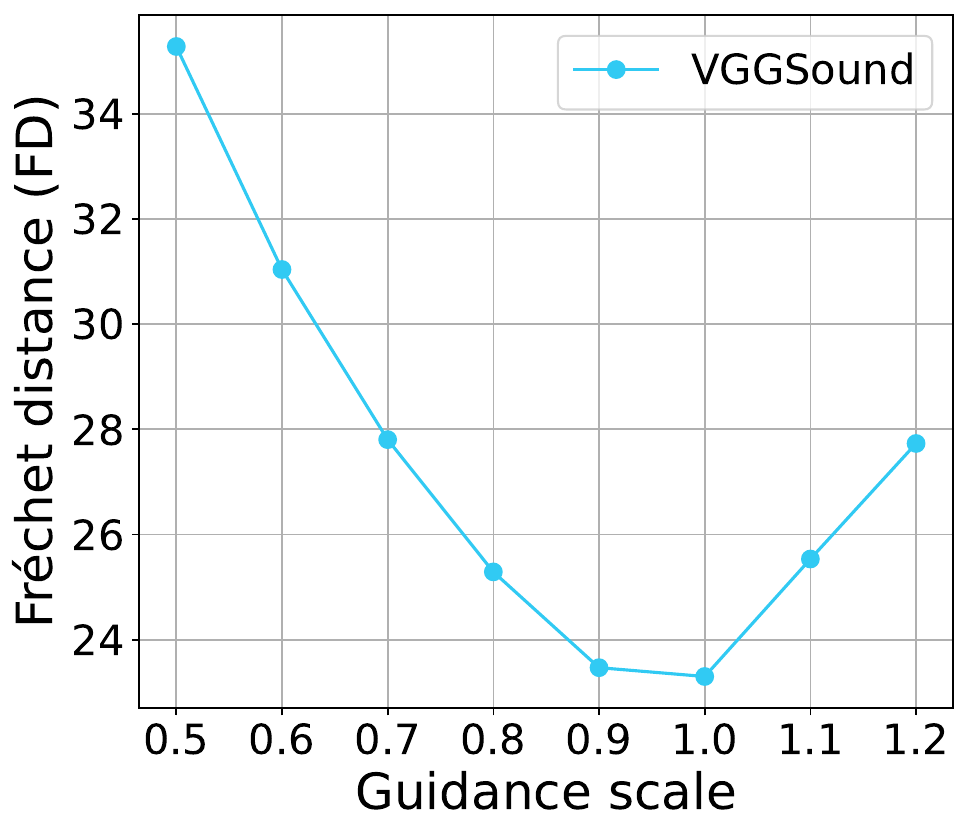}\label{fig:fd-mapper}}
\subfigure[CS.]{\includegraphics[width=0.23\textwidth]{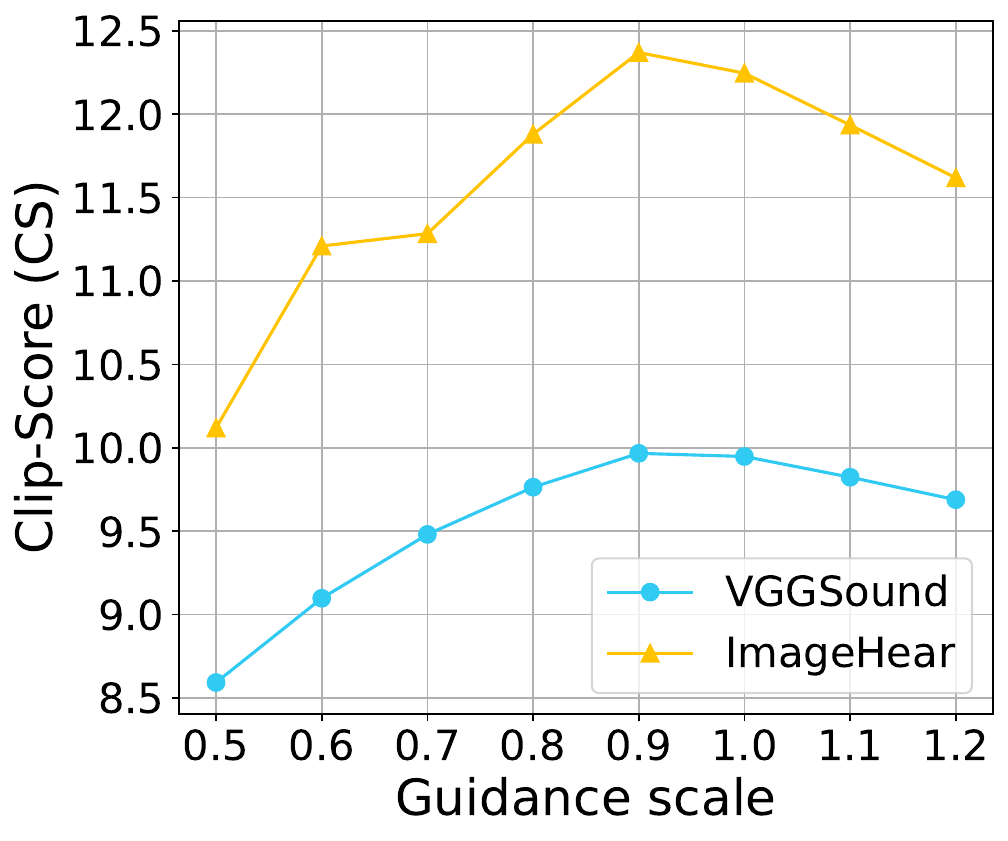}\label{fig:cs-mapper}}
\caption{Ablation study with different guidance scales of the V2A-Mapper.}
\label{fig:gs-mapper}
\end{figure}

\subsubsection{Guidance scale of AudioLDM.}
 Keeping the guidance scale of the V2A-Mapper the same, we run the audio generator with different guidance scales. Fig.~\ref{fig:gs} presents the performance of different guidance scales on fidelity and relevance. As the guidance scale increases, there is a performance gain in the beginning and then a slight drop.

\subsubsection{Guidance scale of the V2A-Mapper.}
When tuning the V2A-Mapper, we keep the guidance scale of AudioLDM as 4.5 as it trades off between the fidelity and relevance. The guidance scale in the V2A-Mapper would directly affect how the visual embedding is translated into the audio embedding. According to the results shown in Fig.~\ref{fig:gs-mapper}, when the guidance scale is 0.9, it has the best performance. We conjecture that relaxing the extent of the condition posed on the diffusion process for the audio embedding generation could make the final sound more diverse and visually relevant.


\begin{figure*}[!t]
\centering     
\includegraphics[width=\textwidth]{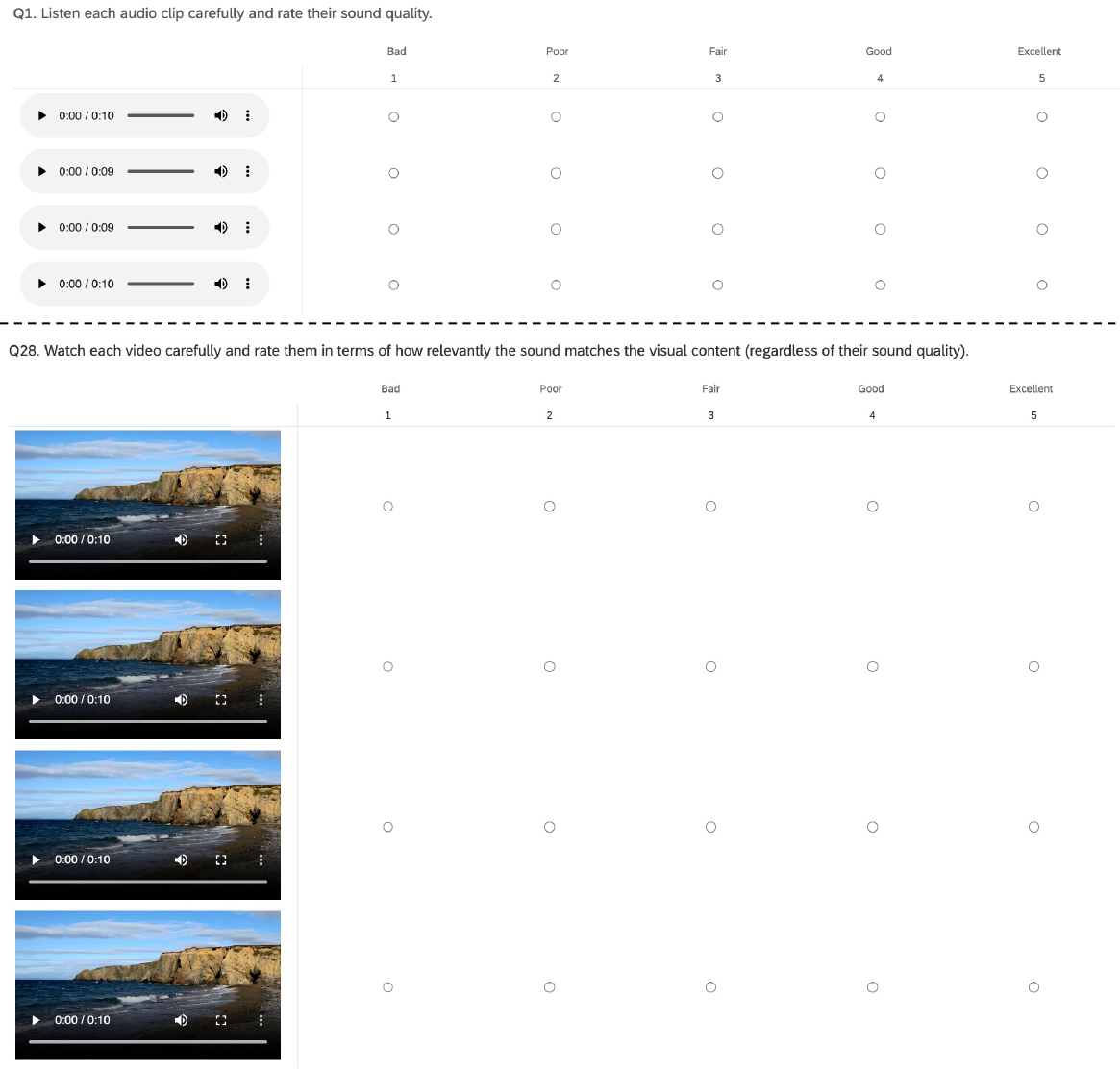}
\caption{\textbf{Top: Example of quality test of video-to-sound generation.} Audio clips from ground truth, previous methods Im2Wav and CLIPSonic-IQ, and our proposed method are presented in random order. Participants are asked to rate them in terms of the sound quality on a scale of 1 to 5. \textbf{Bottom: Example of relevance test of video-to-sound generation.} Videos with audio clips from ground truth, previous methods Im2Wav and CLIPSonic-IQ, and our proposed method are presented in random order. Participants are asked to rate them in terms of the relevance on a scale of 1 to 5.}
\label{fig:subj_vi2sound}
\end{figure*}

\begin{figure*}[!t]
\centering     
\includegraphics[width=\textwidth]{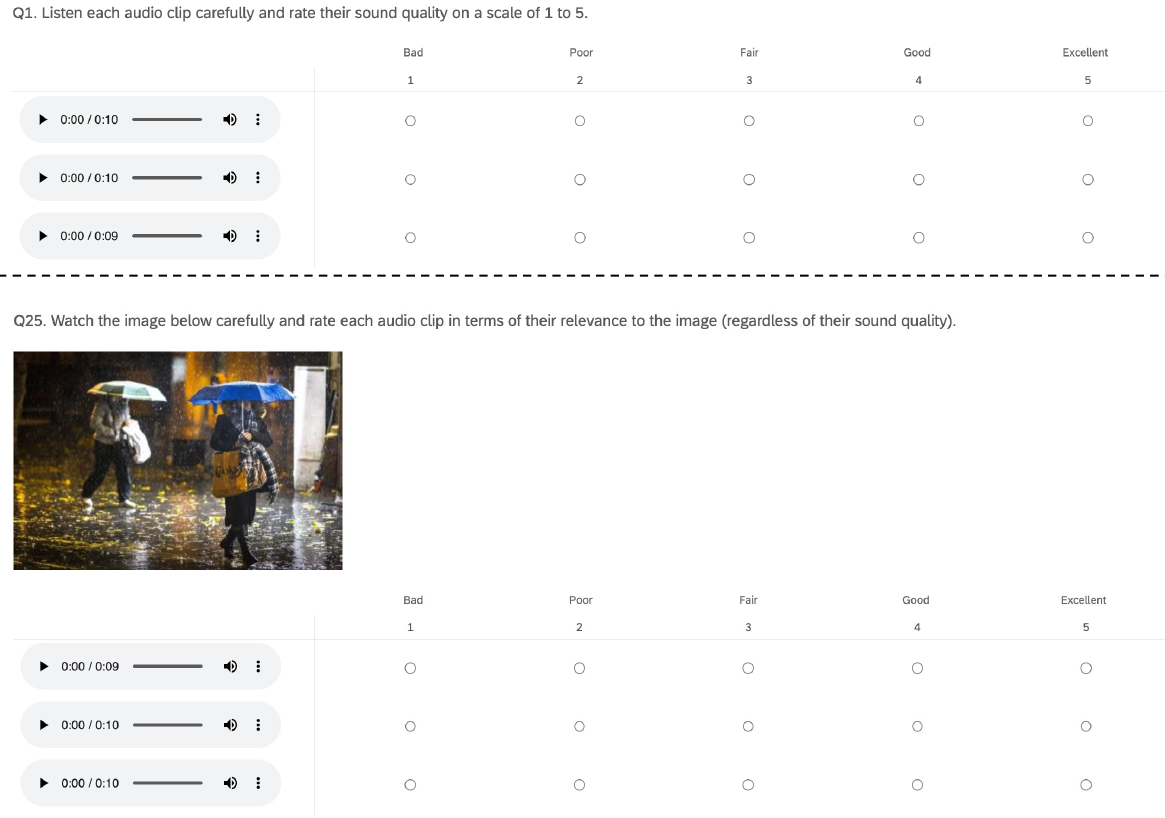}
\caption{\textbf{Top: Example of quality test of image-to-sound generation.} Audio clips from previous methods Im2Wav and CLIPSonic-IQ, and our proposed method are presented in random order. Participants are asked to rate them in terms of the sound quality on a scale of 1 to 5. \textbf{Bottom: Example of relevance test of image-to-sound generation.} Audio clips generated based on the provided image from previous methods Im2Wav and CLIPSonic-IQ, and our proposed method are presented in random order. Participants are asked to rate them in terms of the relevance to the provided image on a scale of 1 to 5.}
\label{fig:subj_im2sound}
\end{figure*}
\section{Human Evaluation}
We conduct two subjective tests - one is for video-to-sound generation and the other is for image-to-sound generation. During each subjective testing, we ask 20 listeners to rate the audio clips of 20 randomly selected visual samples on the dimensions of (1) basic audio quality and (2) relevance of the audio to the visual prompt. For video-to-sound testing, we provide audio clips from the ground truth videos, previous methods Im2Wav~\cite{im2wav} and CLIPSonic-IQ~\cite{clipsonic}, and our proposed method. An example of the video-to-sound generation survey is displayed in Fig.~\ref{fig:subj_vi2sound}. For quality evaluation, participants are asked to rate each audio clips in terms of their sound quality on a scale of 1 to 5. For relevance testing, we replace the audio track of the video with the generated audio and provide four video clips (from ground truth, Im2Wav, CLIPSonic-IQ, and our proposed method) to the participants. Participants are asked to watch each video and rate based on the relevance of the sound with respect to the visual prompt. A similar process is designed for image-to-sound generation survey but no ground truth sound is available. As presented in Fig.~\ref{fig:subj_im2sound}, three audio clips are provided for participants to rate their quality and relevance to the image.

We analyze the collected Absolute Category Ratings (ACR) in two ways. The first analysis, which results in the Mean Opinion Score (MOS), averages the ratings across all listeners separately for each algorithm. This approach captures the information contained in which of the five categories (Excellent, Good, Fair, Poor, or Bad) a listener had selected and is a standard procedure to analyze ACR data~\cite{mos_correct}. However, in our test, listeners made several ACRs at the same time because all the algorithms associated with one visual were presented on one page and listeners were free to listen to all of them before making the ratings. Some listeners reported that they partially or totally ignored the labels on the ACR scale and instead made their selection to reflect their ranking of the algorithms. In response, our second analysis recodes the ACR responses into paired comparisons and infers the relative standings via indirect scaling. Specifically, we code each listener’s ratings by forming all pairwise contrasts between algorithms and counting the number of times one algorithm was rated higher than the other as well as the number of times both were rated the same (ties). We analyze the resulting contingency table to derive scale values for each algorithm~\cite{hannes_rankorder} and normalize these scale values to the smallest perceptual difference needed to break a tie as Just Meaningful Difference (JMD) score. This analysis discards some of the information in the ACR, but, in contrast to our first analysis, retains each listener’s ranking of algorithms. The worth parameters derived with the two methods (MOS for the first and JMD for the second analysis) are highly correlated and 1 JMD corresponds to about 0.5 MOS. Because the two methods of analysis lead to comparable results we feel confident that our test accurately captured the relative worth of the algorithms under test.

\end{appendices}
\end{document}